\documentclass[final]{cvpr}

\usepackage{times}
\usepackage{epsfig}
\usepackage{graphicx}
\usepackage{amsmath}
\usepackage{amssymb}

\usepackage{times}
\usepackage{epsfig}
\usepackage{textcomp}
\usepackage{multirow}
\usepackage{adjustbox}
\usepackage{colortbl}
\usepackage{hhline}
\usepackage{subcaption}
\usepackage{enumitem}
\usepackage{verbatim}
\usepackage{adjustbox}
\usepackage{siunitx}

\usepackage{gensymb}
\usepackage{xcolor}

\usepackage[colorlinks,pagebackref=false,citecolor=blue,bookmarks=false,hypertexnames=true]{hyperref}

\def\cvprPaperID{****} 
\def\confYear{CVPR 2021}

\begin{document}

\title{Trained Trajectory based Automated Parking System using Visual SLAM on Surround View Cameras}	

\author{Nivedita Tripathi and Senthil Yogamani\\
{\normalsize Valeo Vision Systems, Ireland }\\
{\tt\small firstname.lastname@valeo.com}
}

\maketitle

\begin{abstract}
Automated Parking is becoming a standard feature in modern vehicles. Existing parking systems build a local map to be able to plan for maneuvering towards a detected slot. Next generation parking systems have an use case where they build a persistent map of the environment where the car is frequently parked, say for example, home parking or office parking. The pre-built map helps in re-localizing the vehicle better when its trying to park the next time. This is achieved by augmenting the parking system with a Visual SLAM pipeline and the feature is called trained trajectory parking in the automotive industry. In this paper, we discuss the use cases, design and implementation of a trained trajectory automated parking system.  The proposed system is deployed on commercial vehicles and the consumer application is illustrated in \url{https://youtu.be/nRWF5KhyJZU}. The focus of this paper is on the application and the details of vision algorithms are kept at high level. 
\end{abstract}

\section{Introduction}

\begin{figure}[t]
\centering
\includegraphics[width=\columnwidth]{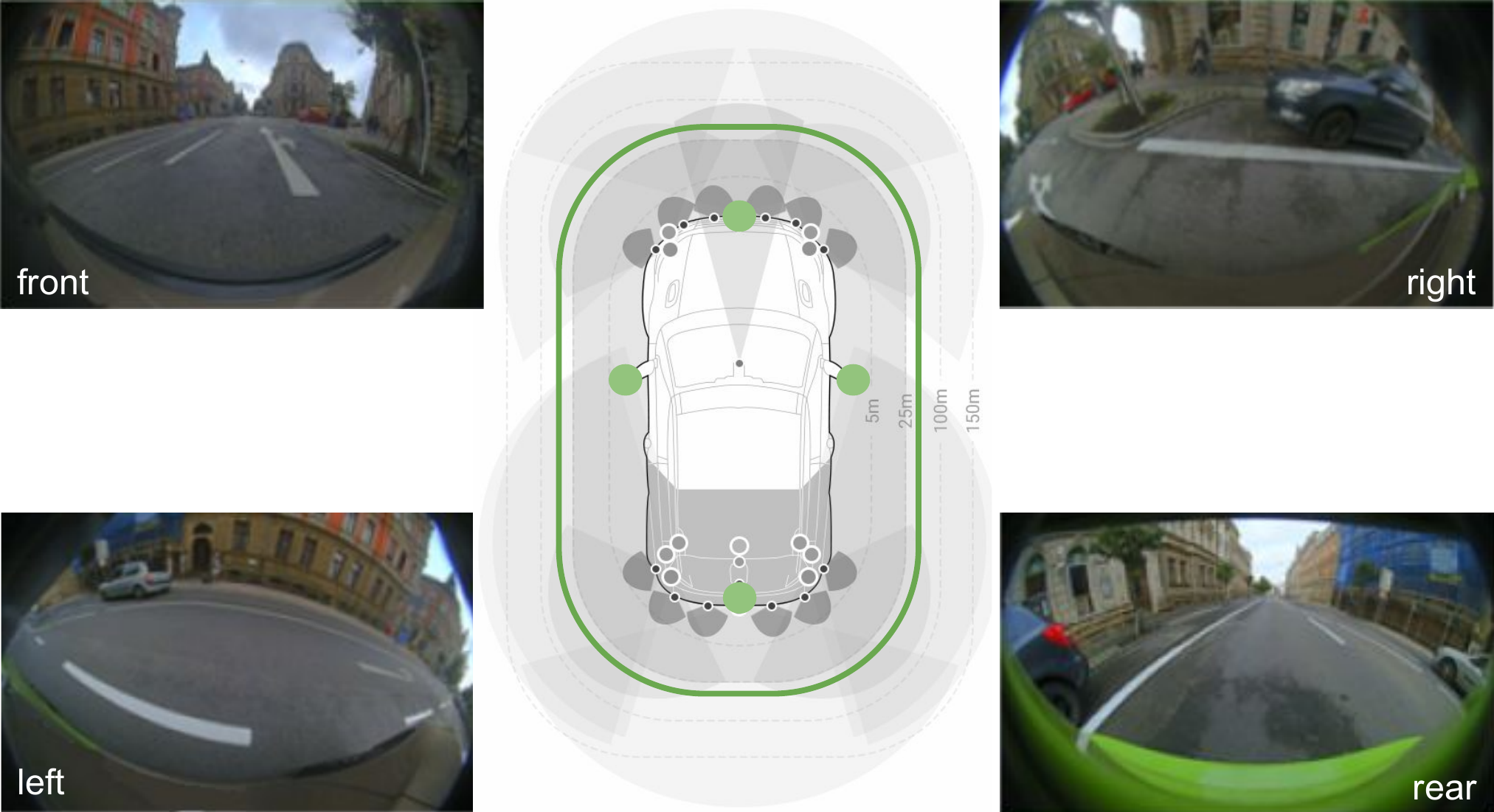}
\caption{Images from the surround-view camera network showing near field sensing and wide field of view. Four fisheye cameras (marked green) provide 360$^\circ$ surround view. }
\label{fig:svs}
\end{figure}

Broadly, Autonomous Driving (AD) use cases can be split into three scenarios according to the speed of operation namely high speed highway driving, medium speed urban driving and low speed parking maneuvers \cite{horgan2015vision}. High speed use cases are relatively well defined and structured and hence features like highway lane keep assist are the most mature and already deployed in the market. Urban driving use cases correspond to medium speed, they are highly unstructured and most challenging. Parking is a low speed use case and it is somewhere in the middle in terms of structuredness. Relatively, the driving rules of parking and its associated road infrastructure (road markings and traffic signs) are less well defined but easier to handle because it is low speed manoeuvring. Parking requires near-field sensing instead of the typical far-field sensing provided by front cameras \cite{heimberger2017computer}. This is typically achieved by four fish-eye cameras which provide full 360$^\circ$ coverage (Figure \ref{fig:svs}) around the near-field of the car.  

It is quite common for a vehicle to repeatedly park in the same areas, e.g: home area of the owner either a garage or in front of the home and office space. An accurate map of the region will aid in automated maneuver to park more efficiently. This can be achieved by means of a Visual SLAM pipeline which builds a map of the parking area which can be used later for re-localization. Typically these parking areas are private regions and are not mapped by commercial mapping companies like TomTom, HERE, etc. Thus the vehicle has to have the intelligence to learn to map frequently parked areas and then relocalize. In this paper, we describe our system which provides this feature using a commercial automotive grade camera and embedded system.

Visual Simultaneous Localization And Mapping (VSLAM) is a well studied problem in robotics and autonomous driving. There are primarily three types of approaches namely (1) Feature based methods, (2) Direct SLAM methods and (3) CNN approaches. Feature based methods make use of descriptive image features for tracking and depth estimation which results in sparse maps. MonoSLAM \cite{Davison:2007:MRS:1263144.1263479}, Parallel Tracking and Mapping (PTAM) \cite{klein07parallel} and ORB-SLAM \cite{mur2015orb} are seminal algorithms of this type. Direct SLAM methods work on the entire image instead of sparse features to aid building a dense map. Dense Tracking and Mapping (DTAM) \cite{Newcombe:2011:DDT:2355573.2356447} and Large-Scale Semi Dense SLAM (LSD-SLAM) \cite{engel2014lsd} are the popular direct methods which are based on minimization of photometric error. CNN based approaches are relatively less mature for Visual SLAM problems and are discussed in detail in \cite{milz2018visual}. Specifically for parking scenarios using surround view fisheye cameras, Visual SLAM was explored in 
\cite{muehlfellner2013evaluation,schwesinger2016automated, qin2020avp}. In general, there is limited work on perception tasks on fisheye cameras but there has been recent progress for tasks such as object detection \cite{rashed2021generalized}, depth estimation \cite{kumar2018monocular}, soiling detection \cite{uricar2021let}, trailer detection \cite{dahal2019deeptrailerassist} and multi-task models \cite{ kumar2021omnidet}.

The rest of the paper is structured as follows. Section \ref{sec:parking} provides an overview of trained trajectory parking system use cases.  Section \ref{sec:sysarch} details the system architecture of a trained trajectory parking system and its components. Section \ref{sec:vslam} discusses Visual SLAM pipeline in detail and its challenges. Section \ref{sec:dataset} details the dataset used and the baseline results. Finally, Section \ref{sec:conc} summarizes the paper and provides potential future directions.

\section{Trained Trajectory Parking System} \label{sec:parking}

\begin{figure}[!t]
\centering
\includegraphics[width=0.48\textwidth]{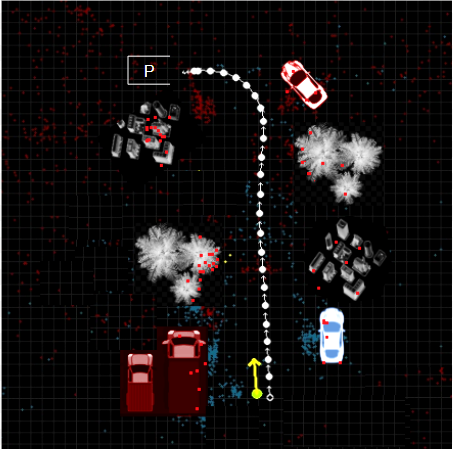}
\caption{Illustration of Trained Parking and Relocalization: The white dotted path is the trained trajectory (with features in red from surrounding objects). Yellow blob with arrow shows the current vehicle (with detected features in blue) moving in direction of arrow, following the trained path.}
\label{fig:reloc}
\end{figure}

Trained trajectory parking works in two phases: training phase and replay phase.
In training phase, a human driver drives the vehicle to park wherever needed (e.g. carport, garage, etc). The trajectory and its other surrounding information are stored for an automated replication at a later time. In replay phase, trained trajectory is loaded to the vehicle and the software is able to recognize the current vehicle's location with respect to the learned trajectory throughout the path. This is illustrated in Figure \ref{fig:reloc}.
Training here refers to the general meaning of this word and  does not adhere specifically to the machine learning terminology. Currently we do not use any machine learning in our VSLAM pipeline. It may cause confusion but this usage is already well established in the automotive industry. Sometimes, it is also called memory parking. In robotics literature, a closely related problem is multi-session mapping and localization.

Fusion of odometry and/or ultrasonic sensors information during training phase gives a more accurate trajectory to replay. Vehicle Control Planning then uses this calculated position of the vehicle to plan a route back to the parking location, and controls the steering and acceleration in order for the vehicle to drive itself there. Visual SLAM algorithm is used for both training and replay phase to calculate and recognize the trained trajectory and vehicle position. These phases of trained trajectory parking are used in different use cases as described below.\\

\begin{figure*}[htpb]
\begin{center}
\includegraphics[width=0.9\textwidth]{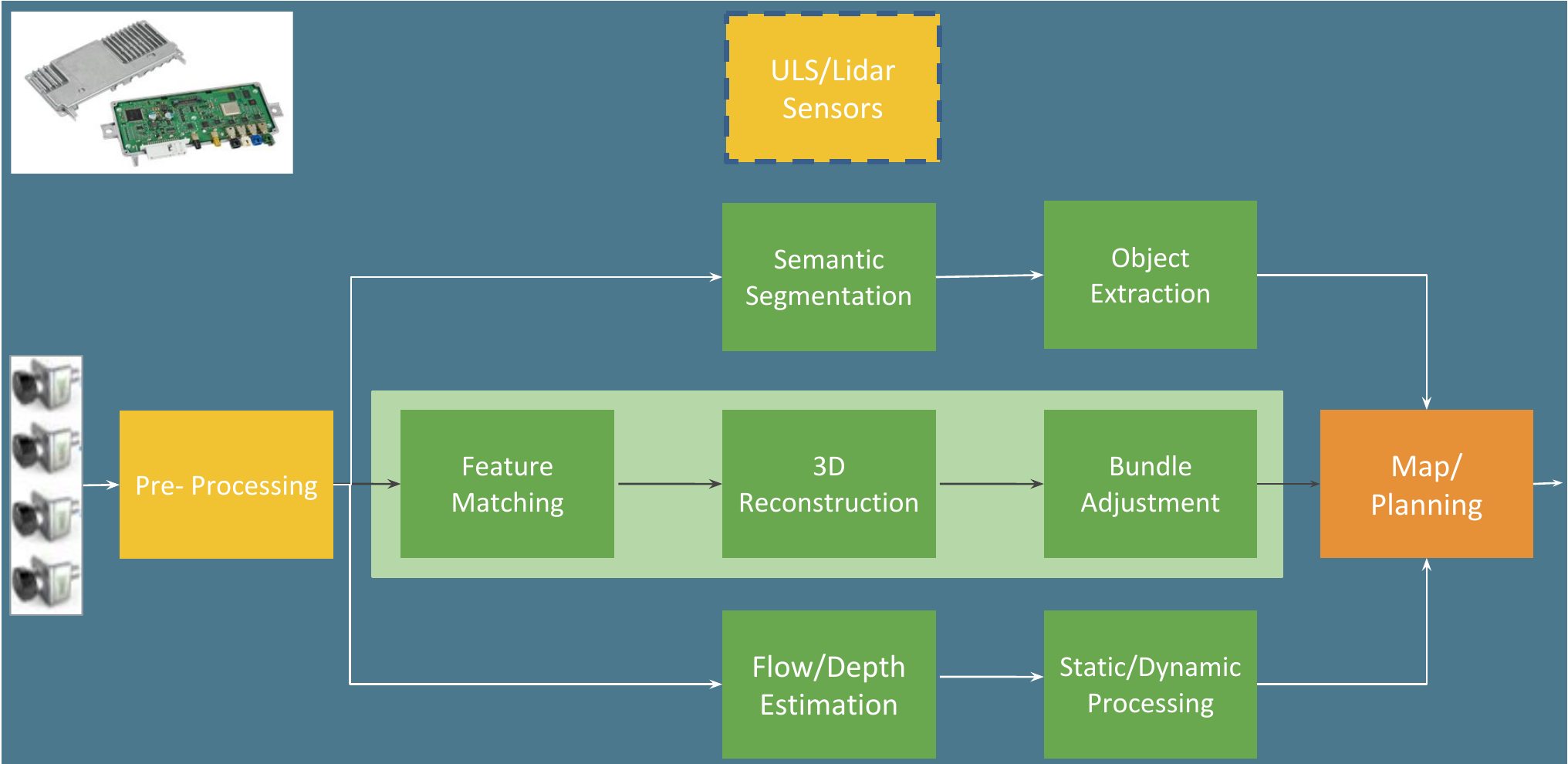}
\end{center}
\caption{Trained Parking System Architecture}
\label{fig:sysarch}
\end{figure*}

\noindent \textbf{Home Parking:} A driver frequently parks the car in their home area and the idea is to learn the home region to automate the parking maneuver.
A home parking system localizes the ego-vehicle using computer vision techniques within an already stored trajectory so that the vehicle is capable of driving completely autonomously into the parking slot using the stored trajectory. In such applications, the driver trains the system to detect landmarks and use it for localization. \\

\noindent \textbf{Automated Reverse Parkout}: This facilitates the driver to reverse any maneuver (e.g. driving into a dead end, parking in). Usually different trained trajectories are stored in buffer on persistent memory, user can then choose the preferred trajectory for automated park-in or park-out based on vehicle's current location. Trajectory for automated replay of park-out gets recorded continuously, generally without any manual trigger. \\ 

\noindent \textbf{Valet Parking:} Valet Parking is the most advanced form of parking which requires Level 4 automation. This system has to autonomously perform navigation to find parking slots, select the optimal one and then park itself. 
It is quite challenging to accomplish this in an unknown environment. Thus there are efforts to create infrastructure maps with artificial landmarks (special QR code like markers) which can be leveraged by vehicles to build an efficient local map. 

\section{Parking System Architecture} \label{sec:sysarch}

The block diagram of our system is illustrated in Figure \ref{fig:sysarch} and described in this section. The necessary computer vision modules required for a parking system are discussed in Section~\ref{subsec:necessary_vision_modules}. Visual SLAM needed for trained trajectory parking is discussed in more detail in Section~\ref{sec:vslam}.

\subsection{Platform Overview}

\textbf{Sensors:} The car setup comprises of commercially deployed automotive grade sensors as shown in Figure \ref{fig:svs}. The primary sensors required for a parking system are fisheye cameras (for providing trajectory information) and Ultrasonic sensors (for proximal obstacle detection on the way to parking). There are four fisheye cameras (marked green in the figure) which are 1 megapixel resolution having a wide horizontal field of view (FOV) of $190^\circ$. The four cameras together cover the entire $360^\circ$ FOV around the car. These cameras are designed to provide optimal near-field sensing upto 10 metres and slightly reduced perception upto 25 metres. There is also an array of 12 Ultrasonic sensors (marked gray in the figure) covering front and rear regions. They provide a robust safety net around the car to avoid collisions which is necessary for a robust system. Typically, an Ultrasonic sensor is composed of a single membrane with modulated pulses for transmission and reception at a  center frequency of \SI{51.2}{\kilo\hertz}. The sensor provides raw data from the piezoelectric element which is followed by an analog filter bank for signal conditional before digital conversion. Further processing steps to detect and localize objects are done on the Electronic Control Unit (ECU). LiDAR is not primarily used for parking as it is focused on far field sensing.\\

\textbf{SOCs:}  Although autonomous driving prototypes are shown on large PCs, they have to be deployed on low-power and low-cost embedded systems. In spite of rapid growth of computational power of automotive embedded systems, it is still quite challenging to deploy computer vision algorithms. Figure \ref{fig:sysarch} shows a typical automotive embedded system called ECU on the top left region. The typical SOC vendors for automotive include Texas Instruments TDAx, Renesas V3H and Nvidia Xavier platforms. Majority of these SOCs provide custom computer vision Hardware accelerators for dense optical flow, stereo disparity and deep learning. A typical SOC system comes with $1$ to $10$ TOPS of computing power and consuming less than $10$ watts of power. \\



\textbf{Software Architecture:} Typical pre-processing algorithms before being fed to vision algorithms includes fisheye distortion correction, contrast enhancement and de-noising. Objects are detected using computer vision algorithms (discussed in section \ref{subsec:necessary_vision_modules}). They are then fed into the map to plan maneuvering for the car for automated parking. The objects in the image coordinates from the four cameras are converted to a centralized co-ordinate system in the world. Depth Estimation provides localization of objects like pedestrians and vehicles detected by semantic segmentation. Similarly, road objects such as lanes and curbs are extracted using connected component algorithm and mapped to world coordinates. Objects can also be extracted from other sensors like Ultrasonics and LiDAR if available which are then fused in the previously used map.

\begin{figure}[t]
\centering
\includegraphics[width=0.48\textwidth]{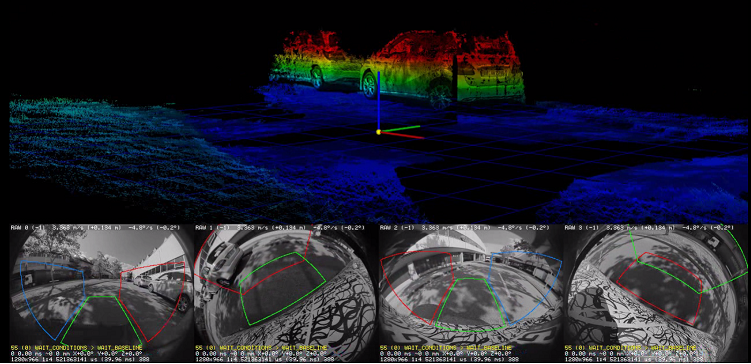}
\caption{Depth estimation via motion stereo}
\label{fig:depth}
\end{figure}

Vehicle Control and Planning unit uses the map and the current position to plan a route back to the parking location, and controls the steering and acceleration in order for the vehicle to drive itself there. GPS can be used to provide a coarse localization of the vehicle at the start of the trajectory. It is also important for the system software to have the ability to detect an obstacle or a pedestrian on the driving path, and change the course of the trajectory accordingly. The system should be able to utilize Automatic Emergency Braking (AEB) functionality to allow vehicle to apply emergency braking under a set of conditions.

\subsection{Standard Computer Vision Modules in Parking} \label{subsec:necessary_vision_modules}

In addition to traditional feature matching, a modern VSLAM system uses semantic information for robustness in re-localization. A modern practice is recognizing the dynamic and movable objects in the scene and give either zero or very little weights to features carried by these entities in the scene.  \\

\textbf{Semantic Segmentation:}
The main objects of interest are road-way objects like freespace, road markings, curbs, etc and dynamic objects like vehicles, pedestrians and cyclists. They can all be detected by a unified semantic segmentation network \cite{siam2017deep} in real-time \cite{siam2018rtseg}. 
These objects are detected in general for navigation and obstacle detection in automated driving. Specifically for our application, dynamic objects can be helpful to eliminate feature points in the map as they may not be in same location during re-localization. Whereas static entities like lane and road markings provide valid trajectories which can be traversed during the maneuver.\\

\begin{figure}[t]
\centering
\includegraphics[width=0.5\textwidth]{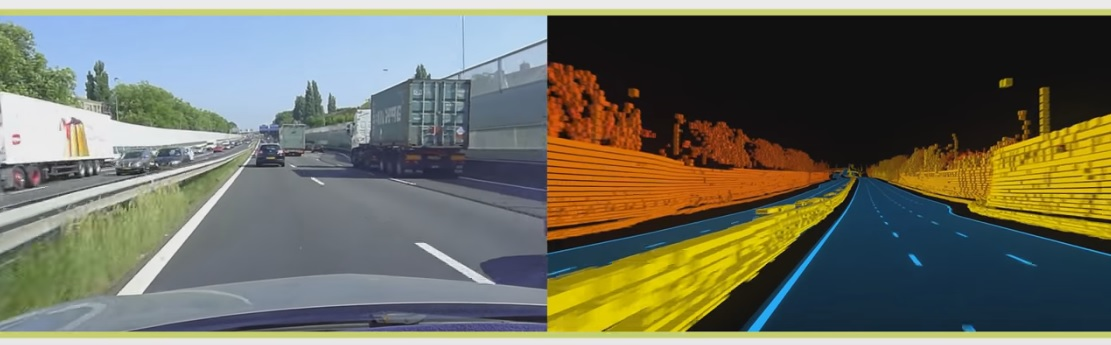}
\caption{Example of High Definition (HD) map from TomTom RoadDNA (Reproduced with permission of the copyright owner)}
\label{fig:HDmaps}
\end{figure}

\begin{figure}[htpb]
\centering
\includegraphics[width=0.45\textwidth]{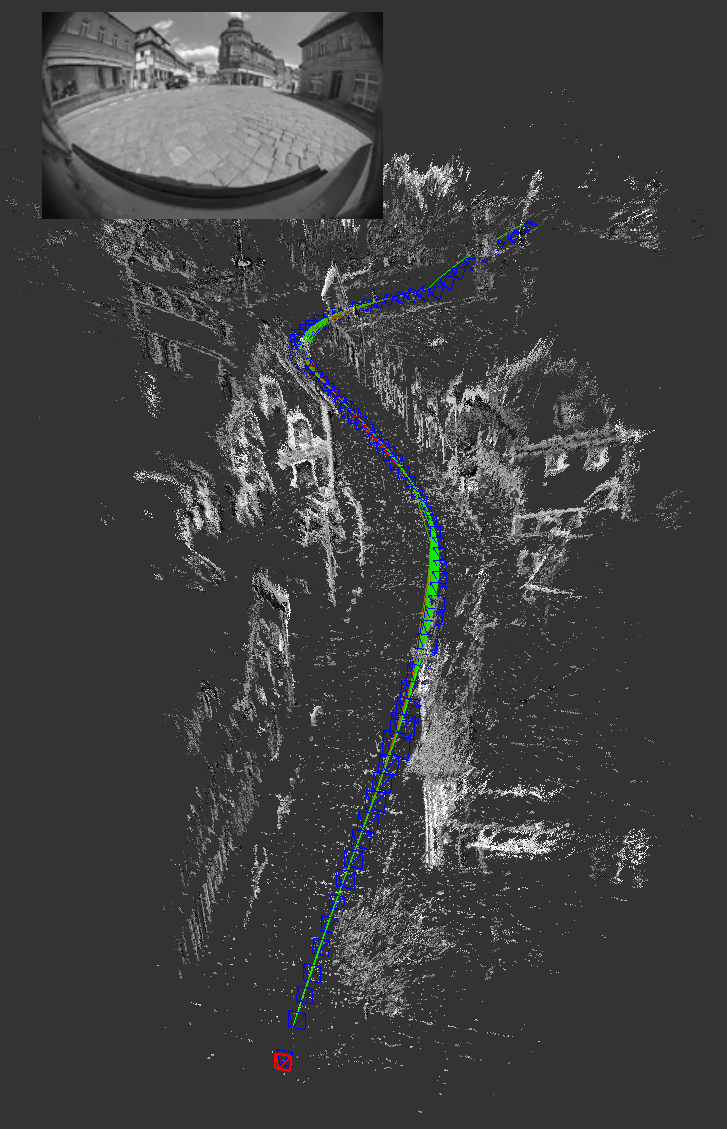}
\caption{Bird's-eye view of camera pose of a trajectory generated by Visual SLAM pipeline along with the corresponding point cloud using motion stereo.}
\label{fig:Pointcloud}
\end{figure}

\begin{figure*}[t]
\centering
\includegraphics[width=\textwidth]{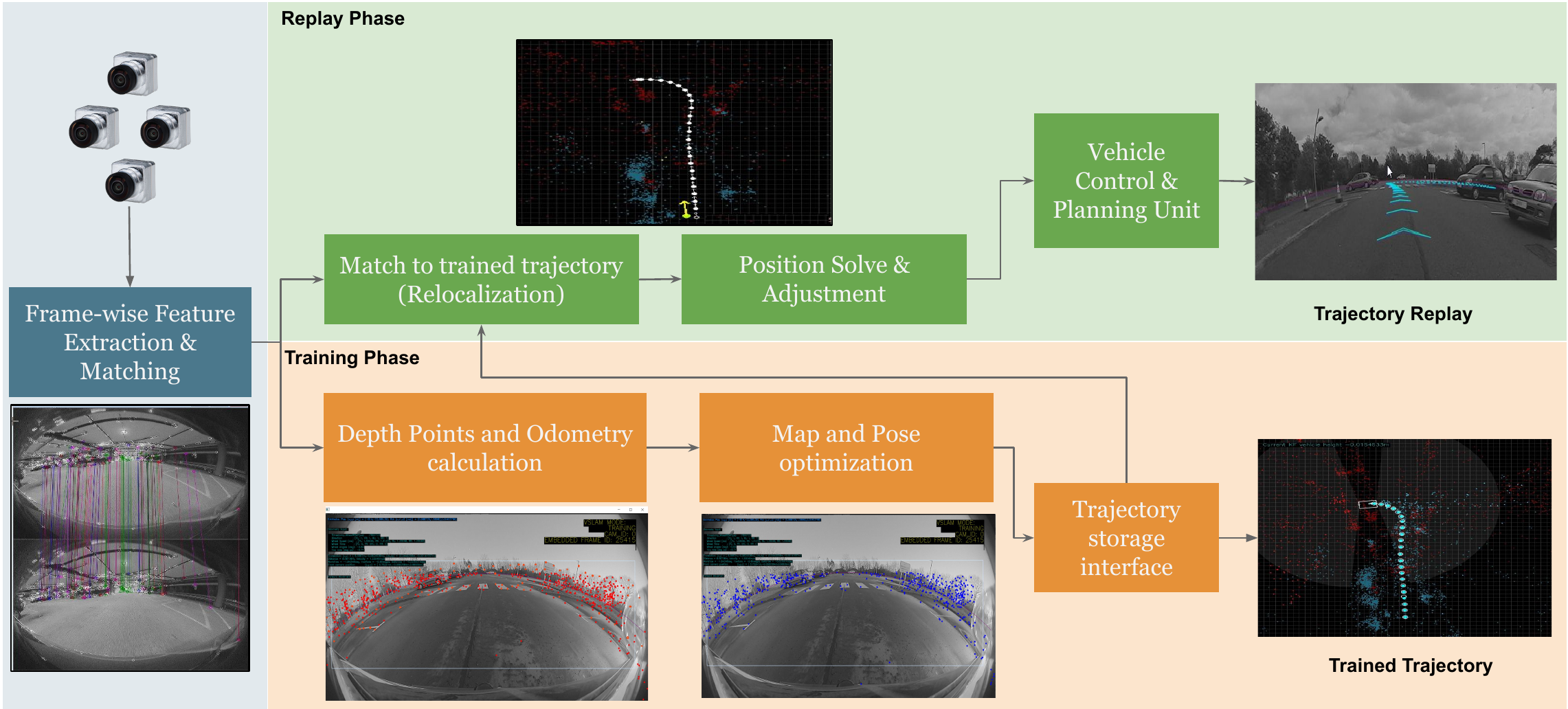}
\caption{Block diagram of VSLAM showing two parallel pipelines for training and replay phases}
\label{fig:VSLAMarch}
\end{figure*}

\textbf{Generic Obstacle Detection:}
In order to obtain a robust system, it is essential to detect objects using alternate cues other than appearance. Training an appearance based semantic segmentation for all possible objects is quite challenging in practice, there are quite rare object classes like Kangaroo or construction truck.  Motion and depth are such cues which are very useful in automotive scenes. Typically, depth is used to detect static objects and motion is used for detecting dynamic objects. As mentioned before, most automotive SOCs provide dense optical flow and stereo hardware accelerators which can be leveraged. The stereo accelerator could be used for motion stereo of our monocular cameras. 
Figure \ref{fig:depth} illustrates depth computed by motion stereo algorithm. In this case, the fisheye distortion manifold is piece-wise planar surfaces which are visualized below the point cloud. Alternatively, they can also be computed using an efficient multi-task network \cite{sistu2019neurall}.


\section{Visual SLAM for Parking}\label{sec:vslam}

\subsection{Mapping Overview}

Mapping is one of the key pillars of autonomous driving. Many first successful demonstrations of autonomous driving (e.g: by Google) were primarily reliant on localization to pre-mapped areas. Figure \ref{fig:HDmaps} illustrates a commercial HD maps service for autonomous driving provided by TomTom RoadDNA \cite{TomTomRoadDNA}. They provide a highly dense semantic 3D point cloud map and localization service for majority of European cities with a typical localization accuracy of 10 cm. When there is an accurate localization, HD maps can be treated as a dominant cue as a strong prior semantic segmentation is already available and it can be refined by an online segmentation algorithm \cite{ravi2018real}. However, this service is expensive as it requires regular maintenance and upgrades of various regions in the world. Due to privacy laws and accessibility, such a commercial service cannot be used in all situations and a mapping mechanism has to be built within a vehicle's embedded system. For example, a private residential area cannot be mapped legally in many countries like Germany. Figure \ref{fig:Pointcloud} demonstrates a point cloud generated by our system. It is quite sparse compared to the dense HD map due to the limited computational power available in a vehicle.

\subsection{VSLAM Pipeline}

Visual Simultaneous Localization And Mapping (VSLAM) is an algorithm that builds a map of the environment surrounding the car, and figures out the current location of the car within that environment, simultaneously. The cameras mounted on the car produce wide angle images from any one or a combination of the four cameras. Then the process of mapping the vehicle's surroundings and tracking the map is followed, which constitutes the basic pipeline of VSLAM visualized in Figure~\ref{fig:VSLAMarch}. 

Mapping is the process of generating a map which consists of a trained trajectory and its associated landmarks, out of the tracked sensor data. A trained trajectory is a group of key poses surrounded by landmarks spanned across vehicle's origin to destination positions. These landmarks are represented using robust image features that are unique in the captured images. On reviewing the state of the art Visual SLAM pipelines, in terms of their advantages and disadvantages, we concluded that a feature based approach would be most suitable over direct methods, as it requires less memory, and is less sensitive to dynamic objects and structure change in the scene. A distinct feature in an image could be a region of pixels where the intensity changes in a particular way, or an edge or a corner. In order to estimate landmarks in the world, {tracking} is performed, wherein two or more views of the same features can be matched. Once the vehicle has moved a sufficient amount, VSLAM takes another image and extracts features. The corresponding features are reconstructed to get their coordinates and poses in real world.

Frame-to-frame 3D reconstruction and visual odometry can have drift and they need to be corrected globally. This is achieved by bundle adjustment step which jointly optimizes 3D points and camera positions. It is a very computationally intensive step as high reprojection errors of 3D points increases the number of iterations to reduce the cost and thus it cannot be performed for every frame. It is typically performed once in N frames and is called as windowed bundle adjustment. At the end of training, full (global) bundle adjustment is also performed wherein all the key frames (not every frame over the trajectory) are optimised to ensure global consistency of internal VSLAM map.

\begin{table*}[t]
\centering
\begin{adjustbox}{width=\textwidth}
\begin{tabular}{|l|l|l|l|l|l|l|l|}
\hline
\multicolumn{3}{|c|}{\textbf{Scene}}                & \multicolumn{2}{c|}{\textbf{Difference}} & \multicolumn{2}{l|}{\textbf{Average Offset}} & \multirow{2}{*}{\begin{tabular}[c]{@{}l@{}}\textbf{Average} \\ \textbf{relocalization  rate}\end{tabular}} \\ \cline{1-7}
\textbf{S.No.}  & \textbf{Training}        & \textbf{Replay}          & \textbf{Time (days)}           & \textbf{Distance (m)}       & \textbf{Position (m)}         & \textbf{Angle (degrees)}             &                                                                                 \\ \hline
Scene1 & 20161208\_121008 & 20161208\_121405 & 0.003          & 4.723          & 0.468            & 4.704         & 81.00\%                                                                              \\ \hline
Scene2 & 20161208\_125225 & 20161208\_125945 & 0.005          & 2.483          & 0.355            & 5.366         & 98.60\%                                                                              \\ \hline
Scene3 & 20161208\_125225 & 20161208\_130048 & 0.006          & 2.692          & 0.3              & 5.149         & 94.30\%                                                                              \\ \hline
Scene4 & 20161208\_125225 & 20170607\_163643 & 181.156        & 2.49           & 1.085            & 8.162         & 74.90\%                                                                              \\ \hline
Scene5 & 20161208\_125319 & 20170607\_163643 & 181.155        & 0.066          & 0.903            & 9.498         & 86.90\%                                                                              \\ \hline
Scene6 & 20161208\_125319 & 20170607\_163529 & 181.154        & 4.96           & 0.896            & 10.751        & 42.80\%                                                                              \\ \hline
\end{tabular}
\end{adjustbox}
\caption{Quantitative results of relocalization rate on selected scenes (higher the relocalization rate, better the performance).}
\label{tab:relocrate}
\end{table*}

The final optimised trajectory gets saved in persistent memory as a map and is used by algorithm to relocalize the vehicle pose for automated maneuvering of the vehicle. During this, the live camera images are searched for features, and are matched to frames from the trained map. If features from the live images are matched to map, optimization module (bundle adjustment) can estimate the current position of the vehicle, relative to where it was during training of the trajectory.

\subsection{Technical Challenges}

We briefly listed below the practical challenges involved in deploying this system based on our experience. \\

\begin{itemize}[nosep]
\item Illumination or weather condition changes can cause the scene to appear visually different. For example, if the mapping and localization are done in day/night or summer/winter etc., the algorithm can degrade significantly as there will be less feature correspondence.
\item Residential areas can have similar structures which makes it difficult for matching unique features. Thus the system needs to be augmented by more specialized features or higher level semantics.
\item Majority of the current generation cars do not have access to cloud infrastructure and thus the mapping has to be done on the car's embedded system. Thus at the end of the trajectory, there is an additional wait time for the driver to allow completion of global bundle adjustment of the map.
\item SLAM pipeline requires good initialization whereby the features along the trajectory can be matched effectively. This is typically done by noisy GPS signal which may cause unreliable relocalization.
\item Structural changes in the scene are quite common due to movement of objects and the map has to be dynamically updated to incorporate these new changes. 
\item Automotive cameras typically have rolling shutter and it has to be compensated especially for relatively higher speeds. 
\item Scale ambiguity is resolved by leveraging metric distance between multiple cameras but there is still possibility of scale drift due to the noise in estimation. 
\end{itemize}

\section{Results} \label{sec:dataset}

\begin{figure}[t]
\centering
\includegraphics[width=0.5\textwidth]{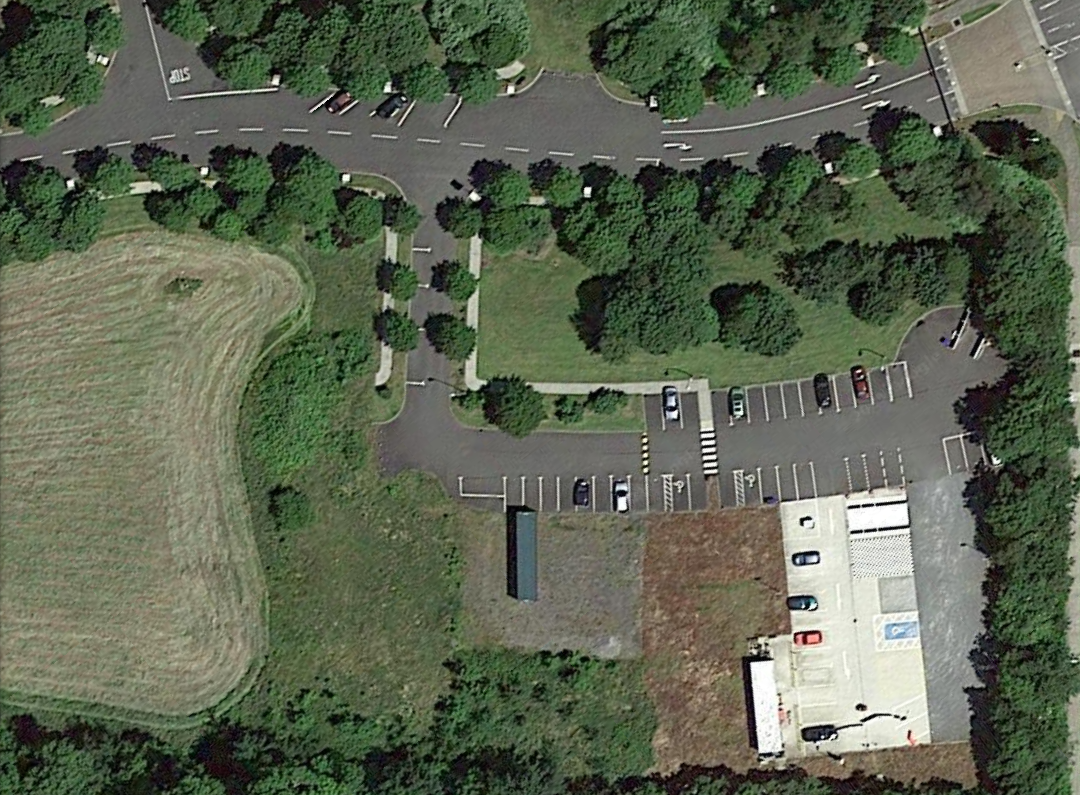}
\caption{Our automated parking test track where the  dataset is captured for evaluation.}
\label{fig:ParkingTestArea}
\end{figure}

The test vehicle consists of four 1 Megapixel RGB fisheye cameras with $190^\circ$ horizontal FOV as shown in Figure \ref{fig:svs} and a Velodyne HDL-64E LiDAR. GNSS(NovAtel Propak6) and IMU sensors(SPAN-IGM-A1) are used to provide ground truth annotation with centimetre level precision.  Vehicle pose with six degrees of freedom obtained for every corresponding image frame is converted into a sequence with some filtering to remove outliers and smooth noise. 
Each element in the set has variations in illumination, weather condition and presence of objects in the scene. The scenes were captured in our test area in Ireland (shown in Figure \ref{fig:ParkingTestArea}) designed for testing various parking scenarios. 

A sample qualitative result of the vehicle relocalizing a previously trained sequence is illustrated in this video \url{https://streamable.com/d6b97}. This sequence illustrates office parking scenario where the car is left at the entrance of a parking lot and it navigates to a designated parking area which was previously traversed. In other sequences, there are home parking like scenarios where the car undergoes a simpler trajectory into a narrow garage parking.
In this video, current front (right) \& reverse (left) view images are shown. The region in the middle shows the trained trajectory map (vehicle poses shown as white dots), with sparse features surrounding it. Moving yellow arrow shows the live movement of vehicle as per the localized positions calculated from the VSLAM algorithm. 


Table \ref{tab:relocrate} presents the results of few selected scenes in our dataset. These scenes have variations in both time/day and lateral/angular offsets causing illumination and structural changes in the video sequences. Relocalization rate gets affected by the amount of variation between training and replay scenes. First three columns in the table refer to training and replay scenes, represented as per the time they were recorded (yymmdd\_hhmmss). Fourth and fifth columns mention the difference of time (in days) at which training and replay scenes were captured, and difference in distance (in meters) between starting of training and replay scenes. Next two columns mention the average offsets of position and angle over the length of trajectory. Last column is average relocalization percentage which is defined as the percentage of instances in which the estimated pose is within a tolerance of 2{\degree} in orientation and 0.05m in position. Scene6 has the most challenging scenario due to large variations in both illumination and lateral offset, thus it has relatively worse relocalization rate.




\section{Conclusion} \label{sec:conc}
In this paper, we provided an overview of an industrial trained trajectory automated parking system. We discussed the trained trajectory parking use cases and demonstrated how to extend current parking systems using a Visual SLAM pipeline. We described the Visual SLAM pipeline in  detail and list the practical challenges encountered in commercial deployment. 
In future work, we plan to explore a unified multi-task network to perform visual SLAM and other object detection modules.





%



\section*{Acknowledgment}
The authors would like to thank our colleagues in VSLAM team for supporting the work and Lucie Yahiaoui for reviewing and providing feedback. We would also like to thank our employer Valeo for encouraging research.

{\small
\bibliographystyle{ieee_fullname}
\bibliography{egbib}
}

\end{document}